# Evolved embodied phase coordination enables robust quadruped robot locomotion

Jørgen Nordmoen
University of Oslo, Norway
jorgehn@ifi.uio.no

Tønnes F. Nygaard
University of Oslo, Norway
tonnesfn@ifi.uio.no

Kai Olav Ellefsen
University of Oslo, Norway
kaiolae@ifi.uio.no

Kyrre Glette
RITMO, University of Oslo, Norway
kyrrehg@ifi.uio.no

## ABSTRACT

Overcoming robotics challenges in the real world requires resilient control systems capable of handling a multitude of environments and unforeseen events. Evolutionary optimization using simulations is a promising way to automatically design such control systems, however, if the disparity between simulation and the real world becomes too large, the optimization process may result in dysfunctional real-world behaviors. In this paper, we address this challenge by considering embodied phase coordination in the evolutionary optimization of a quadruped robot controller based on central pattern generators. With this method, leg phases, and indirectly also inter-leg coordination, are influenced by sensor feedback. By comparing two very similar control systems we gain insight into how the sensory feedback approach affects the evolved parameters of the control system, and how the performances differ in simulation, in transferal to the real world, and to different real-world environments. We show that evolution enables the design of a control system with embodied phase coordination which is more complex than previously seen approaches, and that this system is capable of controlling a real-world multi-jointed quadruped robot. The approach reduces the performance discrepancy between simulation and the real world, and displays robustness towards new environments.

## KEYWORDS

TEGOTAE, CPG, evolutionary robotics

### ACM Reference Format:

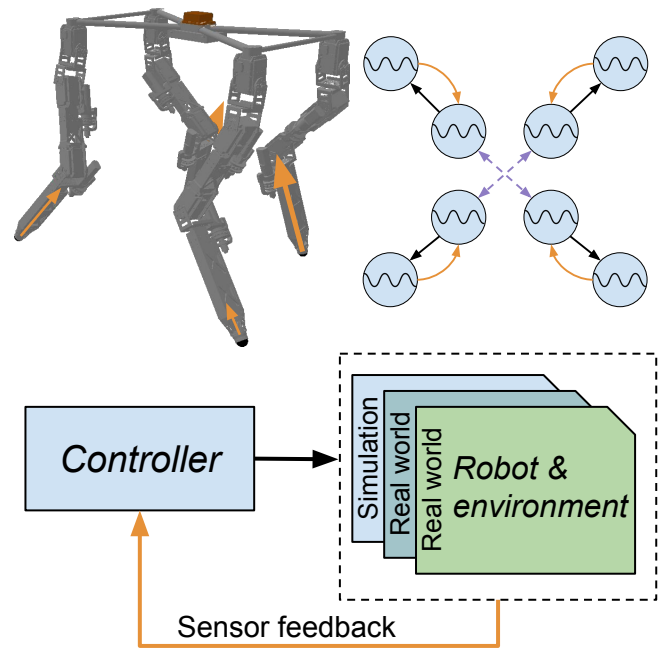

Figure 1: Ground Reaction Force acting on the legs of the robot (shown with orange arrows) aid the Central Pattern Generator (CPG) control system in handling different environments through sensor feedback and embodied phase coordination (illustrated with purple arrows).

## 1 INTRODUCTION

Legged robots are an important means for increasing robot presence in everyday life and can be a valuable tool in difficult tasks such as search and rescue. Because of their increased mobility, they promise to aid on the user's terms instead of requiring the user to accommodate the robot. To achieve this vision legged robots need to be able to adapt to unknown and changing environments, a feat that is made more difficult by the robots own morphology when compared to simpler wheeled robots.

A promising avenue of research is Evolutionary Robotics (ER) which aims to adapt both a robot's shape and its control system to new challenges [7]. In ER adaptation takes place over many trials in a setup which often leverages software simulation to allow complete oversight and reduce experiment time. One of the biggest challenges in ER is the transition from simulation to the real world. The optimization process taking place in simulation, adapting the controller to the robot and the environment, may deviate from useful real-world behaviors [13]. This difference in behavior between the simulated and the real robot is often called *the reality gap* [16].

Many different approaches to tackling the reality gap have been proposed in literature. One possible way to deal with the challenge would be to accept that there are differences between the simulator and the environment, in the same way as the robot will encounter environmental differences in the real world. In other words, the simulation is treated as just another environment that needs controller adaptation [33].

One way to achieve this is to sense the environment and have the control system react based on information gathered from sensors. For complex legged robots this task is difficult because of the need for coordination intralimb and interlimb where each joint might be dependent on disparate sensor inputs [4]. Indeed, the use of sensory feedback for controller adaptation in the ER field has mostly been seen in wheeled robot applications and are less common when legged robots are used as the application domain. Some examples include Morse et al. [15] and Tarapore and Mouret [32] which both utilize touch sensors to trigger an instantaneous phase reset, another example is Gay et al. [8] which combined a Central Pattern Generator with neural network sensor feedback achieving continuous adaptation to sensor input, however, requiring comprehensive engineering of both Central Pattern Generator and the neural network.

An intriguing approach to incorporate sensor feedback in robotics is the TEGOTAE approach [24, 25]. TEGOTAE relies on the concept of embodiment, leveraging the robot body to simplify control [27]. Specifically, TEGOTAE utilizes Ground Reaction Force (GRF) sensors to adapt phase between legs without explicit coordination. The phase of each leg is controlled individually and is slowed- or sped-up depending sensor feedback, implicitly allowing for *embodied control of phase coordination*. In this way, the TEGOTAE approach is less complex than explicit phase coordination because it does not require phase differences to be optimized or specified up front. Because TEGOTAE only affects a small portion of the overall control structure it can easily be combined with different control approaches, widening its appeal for ER research. However, most of the work dealing with TEGOTAE so far has focused on analyzing locomotion to show the advantages of the TEGOTAE sensory feedback mechanism [3], using hand-tuned parameters and inverse kinematics or single jointed legs with low complexity controllers.

In this paper, we demonstrate that we can use ER to incorporate embodied phase coordination for a complex control task: We use the hard-to-balance quadruped 'DyRET' robot, shown in simulation in Figure 1, with a Central Pattern Generator (CPG)-based control system for directly controlling three joints per leg. We evolve two similar CPG control systems, where the first system does not incorporate feedback and the other enables embodied phase coordination through sensor feedback. We compare the simulation results of the two control systems with real-world re-evaluation to understand the effect of sensor feedback on the controllers in the context of the reality gap. Lastly, we perform a case-study where the two control systems are tested in two different and more difficult real-world environments, giving insight into the adaptability of the different controller approaches.

Our results show that embodied phase coordination can readily be combined with a complex CPG control system and achieve well-performing gaits through evolutionary optimization. Where the plain CPG controller achieves high performance in simulation, it suffers from a significant reduction in performance when transferred to the real world. With embodied phase coordination enabled, the control system does not achieve unrealistically high performance in the simulation, and when re-evaluated in the real world the simulated performance is almost fully retained. Stability is better than for the plain controller, and the speed is similar to the plain controller after a warm-up period. Through our case study, we can also observe that sensor feedback makes the control system more robust than the plain controller when transitioning to new and more difficult surfaces.

The contributions of our paper are the following. Firstly we show that embodied phase coordination can readily be combined with an extensive CPG control system on a difficult-to-control quadruped robot. Secondly, we show that the full control system can be evolved and results in robust controllers with little reality gap. In addition, our case study demonstrates how embodied phase coordination allows our control system to transition to unknown environments. Our contributions aid both the understanding of TEGOTAE as a control system mechanism and also as a technique in ER for environmental adaptation and reduced reality gap.

## 2 BACKGROUND

In this section we will review related and relevant work regarding ER and the reality gap, control systems in the field of ER and TEGOTAE.

### 2.1 Evolutionary Robotics and the reality gap

ER draws on principles like selection, variation and hereditary traits found in biological evolution to design robots with embodied intelligence [7]. In the early days of the field experiments were often conducted in the real world, however, this trend has shifted in recent years to favor evolution in simulation. Simulation allows for more control of the environment and rapid verification, however, it also brings with it some challenges. The reality gap is the discrepancy that often occurs between the performance of a robot in software simulation and real-world testing [12]. This problem is challenging in ER since robots evolved in simulation tend to become finely tuned to the simulator including the areas where the simulator disagrees with the real world [16, 29].

The easiest way to avoid the reality gap is to simply not evolve the controller in simulation [19, 21, 23]. This approach avoids the reality gap, but hardware evolution still has its limitations, including the limited number of evaluations due to time constraints, the fact that hardware will wear and deteriorate and the problem of breakage while exploring suboptimal movement during controller adaptation [7]. For these reasons other approaches are still sought after, but always with real-world testing as the final verification [18].

Approaches to solving the reality gap include the introduction of noise in simulation [12], adding obstacles to promote robustness [9], optimizing the simulation to better replicate the real world [36] or ensuring that simulation and the real world agree on the evaluations [13]. Another solution is to include sensor feedback so that the algorithm can adapt online to the current environment [33], however, this approach is seldom used because of the difficulty in integrating sensor feedback and the time-consuming task of calibrating simulated sensors to the real world [29].

## 2.2 Controllers in Evolutionary Robotics

ER has motivated many different control systems for legged robots [4]. Artificial Neural Networks (ANNs) have been used extensively because of their ability to represent complex functions, given the right evolved structure and connection weights [5, 14, 34]. Extensions to these control systems have also come like the SUPG approach [15] which combines the advantages of neuroevolution with sensor feedback. At the other end of the complexity scale are simpler controllers that directly calculate joint angles based on splines or sine waves [6, 20]. These simpler control systems are often used because of the inherent complexity involved in evolving ANNs [31].

Another alternative is the CPG control system [11]. This biologically inspired controller is popular because of the wide diversity in implementation, from a mathematical model [8] to generative encoding [32], and its inherent flexibility in combining phase-coupling with traditional kinematic control [35]. CPG control systems have also been investigated in relation to incorporating sensor feedback [1, 2], however, these approaches require extensive engineering to incorporate sensor feedback which can be unsatisfactory if not the main focus of the intended work.

## 2.3 TEGOTAE

TEGOTAE is a minimalistic approach to utilize sensor feedback for emergent phase-coupling between the legs of the body [25]. Instead of explicitly forcing a given phase-coupling between legs the system relies on GRF sensors and an explicit decoupling of legs to adapt each leg to the environment while simultaneously coordinating the legs through the body. Because TEGOTAE feedback utilizes GRF sensors simulator calibration is limited to accurate weight estimation of the robot and correct physic simulation. This is in contrast to other sensor-based approaches which often require extensive calibration [12].

TEGOTAE has several advantageous properties such as spontaneous gait transitioning [24], robustness to rough terrain [17] in addition to the emergent phase adaptation. Because of the minimal definition, TEGOTAE can also be combined with several different control systems making it an interesting candidate for control system research within the ER community. Most of the work dealing with TEGOTAE has focused on analyzing locomotion to show the advantages of the TEGOTAE sensory feedback mechanism [3], in this paper we extend that research to encompass ER and argue that TEGOTAE sensor feedback is versatile, can readily be combined with a complex CPG control system without the use of inverse kinematics, is robust in light of controller evolution and can work for complex quadruped robots.

## 3 ROBOT AND CONTROL SYSTEM

In this section we will describe the four-legged robot, the control systems and the evolutionary setup utilized for the experiments in the paper[1].

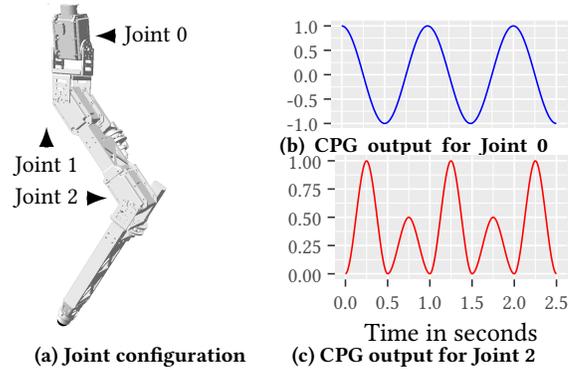

(a) Joint configuration
(b) CPG output for Joint 0
(c) CPG output for Joint 2

Figure 2: (a) shows a visual representation of a leg of the robot with joints marked. (b) shows an example control curve for Joint 0 and Joint 1 while (c) shows an example control curve for Joint 2.

## 3.1 Robot

We use the custom developed 'DyRET' platform [22], shown in simulation in Figure 1, together with Robot Operating System (ROS) [28] and Gazebo/ODE for simulation. 'DyRET' is a four legged, quadruped, robot with a mammalian morphology. Each of the four legs contains three rotation joints as illustrated in Figure 2a. Each of the joints contains PID controllers to which our control system periodically sends desired joint angles. The simulation and the real-world robot operates on the same set of input[2], which allows us to rapidly change between simulation and real-world experiments.

To measure GRF, force sensors of type 'OptoForce OMD-20-SE-40N' are attached to each leg. Simulated versions of the GRF sensors are also utilized during evolution. Of note is that the simulated GRF sensors are not calibrated to the real-world sensors, they work through weight and gravity simulation alone. To measure the pose of the robot we utilize an 'OptiTrack' motion capture system for real world evaluations and direct measurements of the body are used in simulation.

## 3.2 Control system

The control system for our robot is based on a network of oscillators, a CPG [11]. The CPG is based on the work of Gay et al. [8] and variations published in related work [30]. The CPG is optimized for producing a quadruped gait with joint 2 containing a swing and stance phase, as shown in Figure 2c. This is advantageous since it makes the foot trajectory capable of level tracing during ground touch. The two equations used to produce the motion of all three joints, for each leg, are given below. Here we follow the same nomenclature as Gay et al. [8].

$$\dot{a}_{\{0,1\}} = \gamma(\mu_{a_{\{0,1\}}} - a_{\{0,1\}}) \quad (1)$$

$$\dot{o}_{\{0,1\}} = \gamma(\mu_{o_{\{0,1\}}} - o_{\{0,1\}}) \quad (2)$$

$$\dot{\phi}_{\{0,1\}} = 2\pi\omega \quad (3)$$

$$\theta_{\{0,1\}} = a_{\{0,1\}} \cos(F_L(\phi_{\{0,1\}})) + o_{\{0,1\}} \quad (4)$$

---

[1] Additional material and software download see https://folk.uio.no/jorgehn/tegotae/

[2] For more information see: https://github.com/dyret-robot/dyret_documentation/

where $F_L$ is a filter applied on the phase given by

$$F_L(\phi_i) = \begin{cases} \frac{\phi_{2\pi}}{2d} & \text{if } \phi_{2\pi} < 2\pi d \\ \frac{\phi_{2\pi} + 2\pi(1-2d)}{2(1-d)} & \text{otherwise} \end{cases}$$

and $\phi_{2\pi} = \phi_i \mod 2\pi$

To achieve the swing and stance phase Joint 2 utilizes the following equations

$$\dot{a}_{2,1} = \gamma(\mu_{a_{2,1}} - a_{2,1}) \quad (5)$$
$$\dot{a}_{2,2} = \gamma(\mu_{a_{2,2}} - a_{2,2}) \quad (6)$$
$$\dot{o}_2 = \gamma(\mu_{o_2} - o_2) \quad (7)$$
$$\theta_2 = a_2 F_\Gamma(\phi_2) + o_2 \quad (8)$$

with

$$a_2 = \begin{cases} a_{2,1} & \text{if } F_L(\phi_2) < \pi \\ a_{2,2} & \text{otherwise} \end{cases} \quad (9)$$

$$F_\Gamma(\phi_i) = \begin{cases} -16\phi_N^3 + 12\phi_N^2 & \text{if } \phi_N < \frac{1}{2} \\ 16(\phi_N - \frac{1}{2})^3 - 12(\phi_N - \frac{1}{2})^2 + 1 & \text{otherwise} \end{cases} \quad (10)$$

$$\phi_N = 2\left(\frac{F_L(\phi_i)}{2\pi} \mod 0.5\right) \quad (11)$$

$a_i$ represents the amplitude of the $i$-th joint, $o_i$ is the static offset for each joint and $a_{2,1}$ and $a_{2,2}$ is the stance and swing amplitudes for joint 2. For each of these there is a corresponding target, $\mu_i$, which describes the desired value of the variable. $\theta_i$ is the output value of each oscillator and $\omega$ is the frequency. $\gamma$ is a positive gain defining the convergence speed of the oscillator and lastly, $d$ is a virtual duty parameter. Joint 1 utilizes the same equations as joint 0 and both joint 1 and joint 2 are internally connected to joint 0 according to $\phi_n = \phi_{n-1} + \psi_n$, where $\psi_n$ is the desired phase shift between oscillators within a leg. A visual representation of the output of equations 4 and 8 can be seen in Figure 2b and 2c, respectively, where we have set $a_0 = 1.0$, $o_0 = 0.0$, $d = 0.5$, $a_{2,1} = 1.0$, $a_{2,2} = 0.5$ and $o_2 = 0.0$.

For the phase connected open-loop CPG controller joint 0 for all legs are phase-coupled in order to synchronize the legs and achieve a coherent gait. The coupling between oscillator $i$ and $j$ is obtained by the following changes to Equation 3

$$\dot{\phi}_{0_i} = 2\pi\omega + \sum_{j=0, j\neq i}^{n} w_{ij} \sin(\phi_{0_j} - \phi_{0_i} - \varphi_{ij}) \quad (12)$$

where $\varphi_{ij}$ is the desired phase difference between each oscillator and $w_{ij}$ is a positive gain defining the coupling strength.

For the closed-loop controller with TEGOTAE sensor feedback, the static phase-coupling is not utilized and GRF is instead sensed to slow or speed up the phase of the oscillators [25]. The changes to Equation 3 are as follows

$$\dot{\phi}_{0_i} = 2\pi\omega - \alpha N_i \cos(\phi_{0_i}) \quad (13)$$

where $N_i$ is the magnitude of the GRF sensed for leg $i$ and $\alpha$ is the attraction coefficient. The equation will speed up or slow down the phase of all CPGs in the leg, trying to stabilize the body with the leg in stance position when force is detected at the foot sensor.

Table 1: Control parameters for the two control systems. All parameters are shared between the to control systems with identical implications except for attraction coefficient which is only applicable to the closed-loop system (†) while coupling strength and phase difference is only applicable to the open-loop control system (‡).

| Category | Name | Variable | Value range |
|---|---|---|---|
| Global | Frequency | $\omega$ | 0.25 |
| | Gain | $\gamma$ | [0.2, 0.6] |
| | Duty Cycle | $d$ | [0.2, 0.8] |
| | Attraction coefficient† | $\alpha$ | [0.005, 0.1] |
| | Coupling strength‡ | $w$ | [0.1, 2.0] |
| | Phase difference‡ | $\varphi_{ij}$ | 0.0, 0.5, 0.25, 0.75 |
| Joint 0 | Target amplitude | $\mu_{r_0}$ | 0.0 |
| | Target offset | $\mu_{o_0}$ | 0.18 |
| Joint 1 | Target amplitude | $\mu_{r_1}$ | [0.0, 0.3] |
| | Target offset | $\mu_{o_1}$ | [0.36, 1.06] |
| | Phase shift | $\psi_1$ | $2\pi[-0.1, 0.1]$ |
| Joint 2 | Target swing | $\mu_{r_{2,1}}$ | [0.0, 0.7] |
| | Target stance | $\mu_{r_{2,2}}$ | [0.0, 0.7] |
| | Target offset | $\mu_{o_2}$ | [0.85, 1.55] |
| | Phase shift | $\psi_2$ | $2\pi[-0.1, 0.1]$ |

For the rest of this paper we will refer to the CPG controller *without* TEGOTAE sensor feedback as *open-loop* and will refer to the CPG controller *with* TEGOTAE sensor feedback as *closed-loop*, borrowing the semantics from control theory literature [26].

The parameters for the gait are shown in Table 1. To limit the search space for the two control systems we have reduced the number of parameters to only represent the control of one leg. This control is then copied and mirrored for the three other legs. The effect of this restriction is that all legs have the same movement only separated by phase. In turn, this limits the behavior of the robot to, intentionally, only move forwards or backwards. For the open-loop control system, we have additionally forced the phase difference to be a regular walking gait, more specifically a static L-S walk [24], and used a single coupling strength variable, $w_{ij} = w$. These static limitations were put in place to ensure that evolution optimized for gaits that do not put unnecessary strain on the real-world robot.

### 3.3 Evolutionary setup

To evolve the controllers, single-objective Covariance Matrix Adaptation Evolutionary Strategy (CMA-ES) [10] was utilized. The parameters for the evolutionary algorithm are shown in Table 2. For each controller we ran the evolutionary algorithm 20 times to gather statistics about the expected performance.

Due to the morphology, tall and heavy legs and a high center of gravity, our robot is more prone to falling during evolution compared to other robotic platforms usually utilized in ER [18]. Because of this, it is important to include some form of stability measure in the fitness function. For the experiments in this paper we utilized the maximum angular deviation of the body from an upright pose as a stability measure. This measure allows for small rapid movements of the body and will act as a force to minimize large angles that are, from experience, often a precursor to falling. To further discourage falling behavior we used distance walked as a fitness measure since it will tend to favor stable walking patterns over quick sprint and fall behavior often experienced when using speed as a measure. Straight-line distance is also advantageous since it should promote gaits that walk in a straight line not turning or doubling back on itself. We compose distance and stability in such a way as to favor distance with stability as an additional reward. The fitness function is given below:

$$F = F_{distance} (1 + F_{stability}) \quad (14)$$

where

$$F_{distance} = dir \, ||P_{end} - P_{start}|| \quad (15)$$

$$F_{stability} = \begin{cases} 1 - \frac{\arg\max_t ||\angle_{Z_t}||}{\angle_{max}} & \text{if } ||\angle_{Z_t}|| < \angle_{max} \\ 0 & \text{otherwise} \end{cases} \quad (16)$$

Table 2: Parameters for the evolutionary algorithm.

| Name | Value |
|---|---|
| Algorithm | CMA-ES |
| Repetitions | 20 |
| Evaluations | 2510 (250 generations) |
| Genome | Real-valued [0, 1] With 10 parameters |
| Evaluation time | 20 seconds |
| $\lambda$ | 10 |
| $N$ | 10 |
| $X$ | 0.5 |
| $\sigma_{initial}$ | 0.3 |
| $\angle_{max}$ | 0.35 radians |

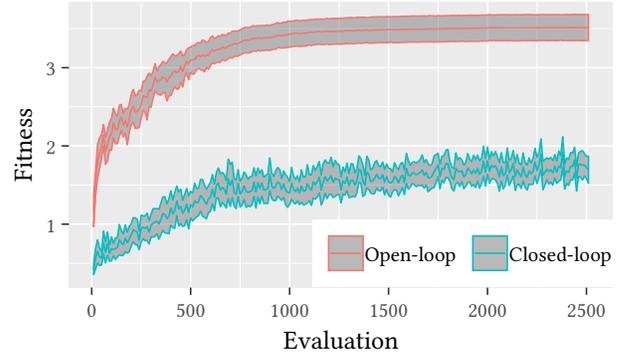

Figure 3: Mean fitness, and 95% confidence interval, for the population best individual across 20 evolutionary runs.

$P_{start}$ and $P_{end}$ is the position of the robot at the start and end of the evaluation, $dir$ is either $-1$ or $1$ dictating the direction of travel, to evolve gaits that walk forward along the $Y-axis$ and $\angle_{Z_t}$ is the angle between the world up vector $\vec{Z}$ and the up vector of the robot pose at time $t$. The angle is normalized to $\angle_{max}$ which is used as the maximum allowable angle deviation, see Table 2 for the specific value used.

## 4 EXPERIMENTS AND RESULTS

The experiments in this paper are focused on evolving gait controllers in simulation for a four-legged robot before evaluating the controllers in the real world. During evolution, we compare solutions for their capability to walk continuously through the composed single-objective fitness measure described in the previous section, Equation 14. We first compare the fitness of the two different control systems throughout evolution to evaluate if they are capable of generating continuous gaits for the whole evaluation period without falling. From the evolutionary runs, we select 5 gait controllers from each control system which will be re-evaluated in software and tested in the real world. Lastly, we perform a case study of two selected controllers in two different environments. By testing the control systems in different environments we can assess how robust the evolved behavior is to external changes and assess the behavioral difference between distinct environments.

### 4.1 Evolutionary run

Figure 3 shows, for both control systems, the mean fitness of the population best controller across 20 evolutionary runs. From the figure it is clear that the open-loop control system has converged while the closed-loop control system displays more variation. Part of the reason for this variation seems to be individual controller variability where repetitions of the same controller or very similar controllers can display noticeable differences in fitness. During evolution, the open-loop control system is capable of attaining much better fitness compared with the closed-loop system and it is also interesting to note that the open-loop system walks further from the beginning of evolution.

To better understand the evolution of parameters for the two control systems we have plotted the whole genome of each control system over time in Figure 4. From the two graphs, we can see that CMA-ES is able to search the whole parameter space initially indicating early exploration of the search space. It is also clear that for the open-loop control system evolution converges to a small range of values for most variables. This is in contrast to the closed-loop control system which has a wider distribution of values for many variables, which converge for the open-loop system. Of note is that the target amplitude of joint 1 ($\mu_{r_1}$) converges to the maximum value for both control systems. This could indicate that evolution would utilize a larger value of the variable. Tests confirm that this does give longer distance, however, at the cost of movement unsuitable for the real-world robot – justifying the restriction. Another interesting observation is that some variables, e.g. $\mu_{o_1}$ and $\mu_{o_2}$, converge to similar values for both control systems, while others, e.g. $\mu_{r_{2,2}}$, have diverged to different values for each control system.

### 4.2 Re-evaluation

To better understand how the evolved control systems behave across the reality gap the 5 best solutions, for each control system, were selected from the last evaluation for re-evaluation in simulation and real-world tests. Re-evaluation in software is done to get an impression of the controller robustness. Note that the variation observed in simulation is due to timing and dynamics of the ROS/Gazebo setup. The following experiments are carried out with 10 repetitions per individual controller, in total 100 real-world evaluations.

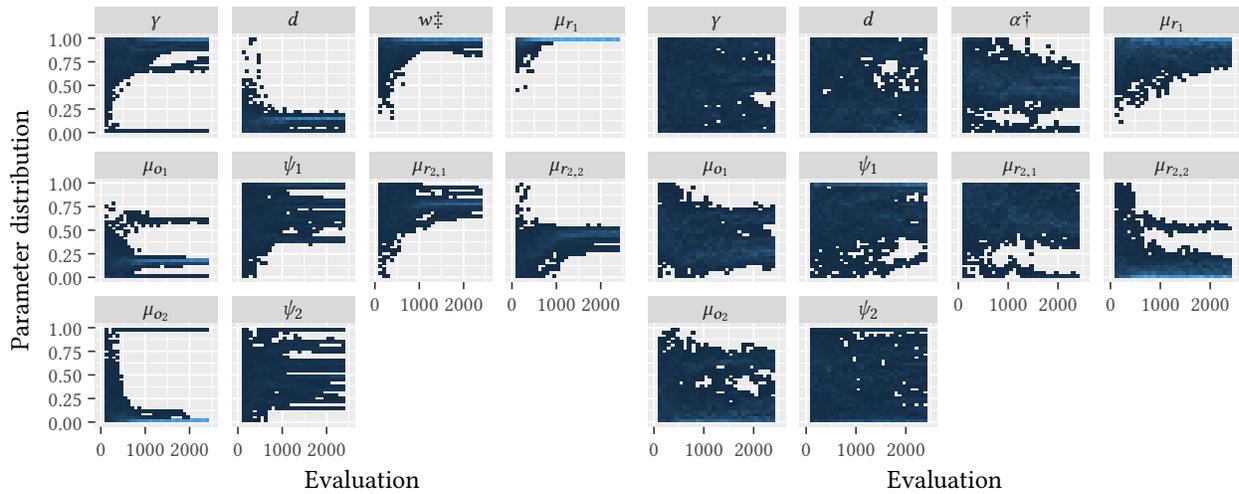

Figure 4: Genotype value distribution throughout evolution of all 20 repetitions, (a) shows parameters for the open-loop (‡) controllers and (b) shows parameters for the closed-loop (†) control system. Note that $w$ and $\alpha$ are unique for each control system and all other variables have the same interpretation.

In Figure 5a the fitness is shown for both control systems. The plot shows, on the left, the fitness after re-evaluation in simulation and, on the right is, the fitness after real-world evaluation[3]. From the plot we can see that the open-loop control system is able to attain better fitness both in simulation and in real-world evaluations. For the closed-loop control system there is more individual variance in simulation compared to real-world evaluations.

Since the fitness measure is composed of two different metrics it can be interesting to separate them out and see how the two control systems differ on each. Figure 5b shows the distance component while Figure 5c shows stability. For distance we can see more or less the same performance difference as with fitness, the open-loop system achieves longer distances both in simulation and in the real world. With regards to stability, the difference between the two controllers is smaller than for distance and the closed-loop system performs better than the open-loop control system in the real world. In contrast to distance, the individual variation is lower for the closed-loop system both in simulation and in real-world evaluations.

One reason why the open-loop controller achieves longer distances compared to the closed-loop control system is the TEGOTAE system apparent need for a warm-up period before it begins to walk. This period is used to build a phase difference between the legs and can be seen in both simulated and real-world evaluations. To gauge the effect of this warm-up period we have plotted the distance traveled in the last half of the evaluation in Figure 5d. From the graph we can see that the difference in simulation is quite considerable, however, in the real world the two control systems perform equally well.

To test if the differences in previous results, Figure 5a, 5b, 5c and 5d, are significant we performed a Mann-Whitney U test for each of the four combinations - comparing both control systems in Simulation and the Real World and comparing Simulation with the Real World for both control systems - for each of the four performance metrics. We select a threshold of significance of $\alpha = \frac{0.05}{16} \approx 0.003$ using Bonferroni correction with 16 comparisons. In summary, every combination except open-loop and closed-loop in 'Real World' for 'Distance - last half' shown in Figure 5d ($\alpha = 0.1991$), are below the selected threshold representing statistical significant differences.

### 4.3 Case study

As a case study, we selected the best controller for each control system, based on fitness, from real-world testing to test in new environments. By understanding how the controllers perform in varied real-world environments we can further characterize the transferability of the embodied phase coordination mechanism in addition to gaining insight into the robustness of our controllers. To create challenging terrains within the confines of our laboratory setup we added two carpets with different characteristics. The first one is a rough hard carpet with large woven knots[4] simulating rough gravel while the other is a soft thick pile carpet with high friction and a more sand like texture[5].

For the case study, we focus on stability and distance in the last half of the evaluation which should show minimal impact from the TEGOTAE warm-up. In Figure 6, on top, the distance in the last half of the evaluation is plotted for each controller for each of the 10 repetitions across all environments tested. From the figure, the performance of both controllers is about equal in the three real-world environments and performance decreases as the environmental difficulty increases.

---

[3]Videos of real-world evaluations: https://folk.uio.no/jorgehn/tegotae/video/

[4]https://folk.uio.no/jorgehn/tegotae/environments/#rough-surface
[5]https://folk.uio.no/jorgehn/tegotae/environments/#soft-surface

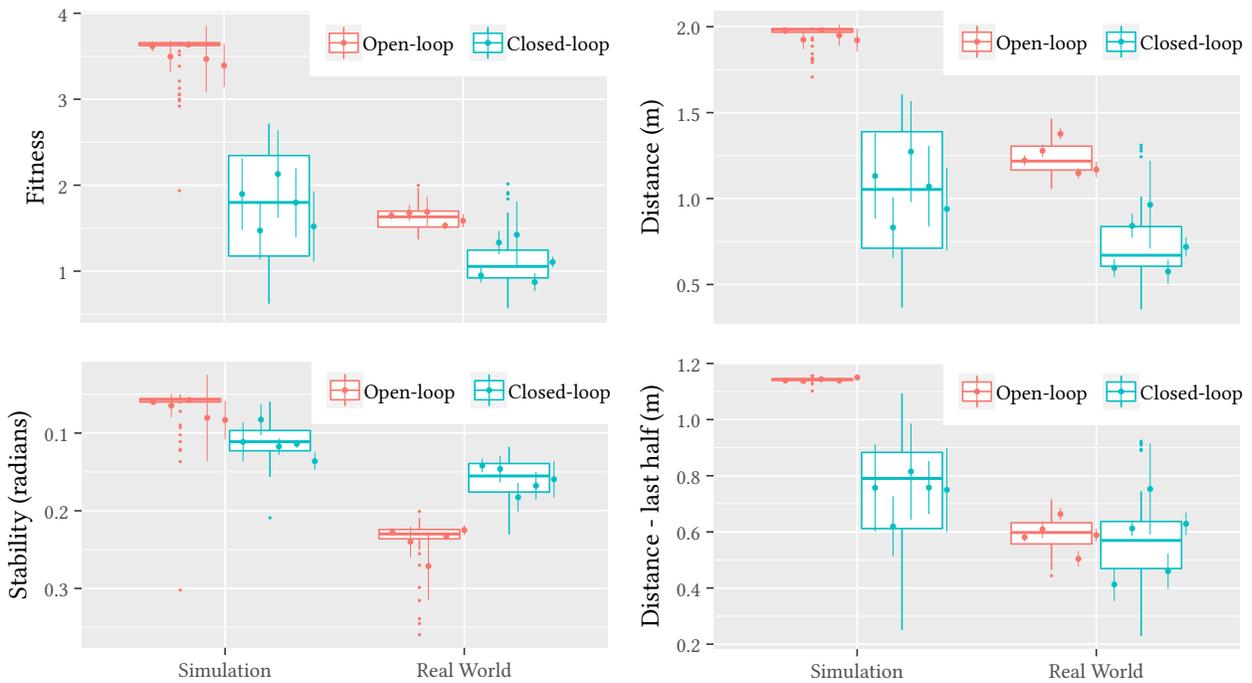

(c) Maximum deviated angle over the whole evaluation. Note the reversed Y-axis since $0.0$ represents perfect stability.

(d) Distance traversed for the last half of the evaluation period.

Figure 5: Data from re-evaluation in simulation and real-world evaluations. The boxplots show aggregate data over all individuals while the mean of each individual controller is shown as larger circles along with $95\%$ confidence intervals.

Stability, shown at the bottom of Figure 6, shows a slightly different trend compared to distance. For both controllers performance decreases when going from simulation to real-world, however, the two additional real-world environments do not see decreasing performance. If we see the result in relation to distance traversed, on top in Figure 6, it is still clear that the different surfaces are difficult to traverse as stability remains almost the same while the distance is halved for the most difficult environment.

We performed the same Mann-Whitney U statistical comparison as before, this time only between the open-loop control system and the closed-loop system, with $\alpha = \frac{0.05}{8} \approx 0.006$. Most of the comparisons show statistically significant differences with the exception of the comparison between the two systems for 'Rough surface' and 'Soft surface' on the 'Distance - last half' metric shown in the top right of Figure 6.

## 5 DISCUSSION

The difference in performance between simulation and real-world experiments, shown in Figure 5a, illustrates that the reality gap is present in our experiments. For both systems the performance decreases, mainly due to lower walking speed, shown in Figure 5b, but also lower stability. The open-loop control system still seems to outperform the closed-loop system after the transition, however, the difference is much smaller and in regards to stability, Figure 5c, the sensor feedback seems to be able to overcome some of the challenges encountered in the real world and outperform the open-loop controller. The main source for the reality gap in both systems is most likely due to simulation inaccuracies as the simulator is not able to model the slight bending of real-world materials nor the umbilical cord and free-hanging wires needed for the real-world robot. However, as the stability results show in Figure 5c, the closed-loop control system seems to adapt to the changes in environmental circumstances emphasizing the advantage of sensor feedback. Interestingly for the closed-loop system variance is reduced between simulation and 'Real World' for the distance metric, again pointing to the gap between reality and simulation.

Because the closed-loop system requires some time to adapt the phase difference between legs it is also interesting to compare distance in the latter half of the evaluation, shown in Figure 5d. The result shows that the real-world performance of the two control systems is similar once TEGOTAE has adapted the phase differences, as they are both able to achieve the same walking speed. This result further illustrates the adaptiveness of the closed-loop control system as it is able to retain much higher stability while at the same time being able to walk with the same speed as the open-loop control system. It should also be noted that the difference between simulation and real-world testing for the closed-loop control system is much lower than for the open-loop system. This indicates that the simulation results for the closed-loop system are more realistic and in turn giving a reduction in the reality gap.

With regard to the case study performed we can see in Figure 6 that distance covered decreases for both systems, indicating that the rough and soft real-world surfaces are more difficult to traverse for both systems, as is evident in the large increase in variance compared to the 'Real World' environment. The graphs also illustrates that the two control systems behave differently through these environmental transitions and that the reduction in performance for the closed-loop system seems to be smaller compared to the open-loop control system. This difference points to the ability of the TEGOTAE sensor feedback to adapt to the environment giving increased robustness. Because of the limited sample size of the case study further comparisons between the two control systems are difficult to validate. Another observation made during the case study is the heavy strain the open-loop system put on the robot joints in the more difficult environments. During the evaluations, both control systems would sometimes get stuck, unable to move one or more legs, explaining the larger variance in Figure 6. For the closed-loop system, this presented less of a problem since sensor feedback would detect the additional leg load and slow movement thus avoiding excessive load on the joints.

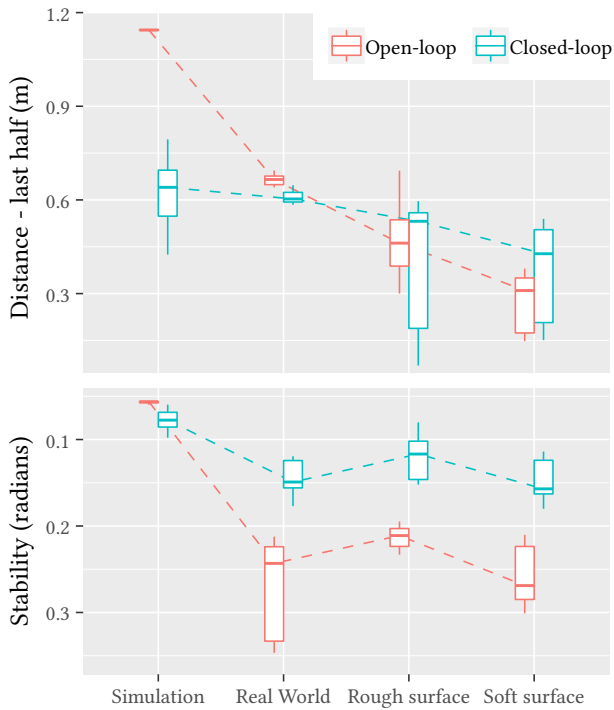

Figure 6: The performance of the two case study controllers for the metrics 'Distance - last half' and 'Stability' for all environments tested. The boxes summarize all 10 evaluations and the dashed lines illustrate the trend of the median. The order is based on median performance on 'Distance - last half' corresponding with the difficulty of the environment.

Because of the similarity of the two control systems, it is interesting to compare the evolution of control parameters, as shown in Figure 4. One apparent property of the graph is that the open-loop system seems to have converged for a large number of parameters compared to the closed-loop system. In light of the performance of the two systems both in simulation, but more importantly in real-world tests, one interpretation of the parameter convergence can be that the open-loop control system has overfitted to simulation. By having sensor feedback in the closed-loop control system, it must adapt to the GRF sensors and is not able to overfit to the perfect conditions of the simulation. This could be the reason for the more robust controllers observed in the transition from simulation to real-world tests. The effect could be related to the technique of adding noise in the simulation to reduce the reality gap [12]. Another interpretation is that because of sensor feedback the closed-loop control system needs longer time to converge and is still in the process of converging. We are planning to address these two interpretations in future studies by including noise during evolution hopefully mitigating the problem of overfitting.

## 6 CONCLUSION AND FUTURE WORK

In this paper we investigated how embodied phase coordination through sensor feedback would affect the evolution, performance and robustness of a CPG control system. We demonstrated that the addition of embodied phase coordination allows the robot to adapt to its environment and is able to produce continuous coordinated gaits for a complex quadrupedal robot. The reduced difference in performance between simulation and the real world, in addition to robustness to new environments, allows for increased confidence in simulation results. Because TEGOTAE sensor feedback can easily be implemented in physics simulators, requiring no sensor calibration, it can efficiently be implemented in other ER research.

Because of the ease of which TEGOTAE can be integrated with complex CPG control systems future research should look into the possibility of first evolving the CPG and later adding sensor feedback to the same controller. This would shed light on the differences in parameter convergence observed in this paper. Additionally, integrating TEGOTAE sensor feedback with a completely different control system should also be attempted to broaden the applicability of the embodied phase coordination mechanism. Another topic to investigate is how to reduce the phase adaptation period observed for the TEGOTAE approach. Ideally the adaptation should occur over a minimal timespan avoiding the need for a 'warm-up' period and maximizing the distance traversed. Since the phase coordination mechanism is dependent on the body of the robot it could also be interesting to discover how the TEGOTAE system would handle changes to the body. A change involving the morphology of the robot would be an interesting experiment along with changes in the body characteristics such as joint velocity.

## ACKNOWLEDGMENTS

This work is partially supported by The Research Council of Norway under grant agreement 240862 and through its Centers of Excellence scheme, project number 262762. The simulations were performed on resources provided by UNINETT Sigma2.